\documentclass[a4paper]{llncs}

\usepackage{graphicx}
\usepackage{subcaption}
\usepackage{amssymb}
\usepackage[utf8]{inputenc}
\usepackage{url}
\usepackage{color}
\usepackage{amsmath}
\usepackage{todonotes}
\usepackage{lipsum}  
\usepackage{xargs}          
\usepackage{tablefootnote}
\usepackage{notoccite}
\usepackage{multirow}
\usepackage{arydshln}

\usepackage[acronym]{glossaries}

\makeglossaries

\newacronym{AutoML}{AutoML}{Automated Machine Learning}
\newacronym{ML}{ML}{Machine Learning}
\newacronym{AI}{AI}{Artificial Intelligence}
\newacronym{NE}{NE}{NeuroEvolution}
\newacronym{ANN}{ANN}{Artificial Neural Network}
\newacronym{DANN}{DANN}{Deep Artificial Neural Network}
\newacronym{EC}{EC}{Evolutionary Computation}
\newacronym{Fast-DENSER}{Fast-DENSER}{Fast Deep Evolutionary Network Structured Representation}
\newacronym{DENSER}{DENSER}{Deep Evolutionary Network Structured Representation}
\newacronym{BNF}{BNF}{Backus-Naur Form}
\newacronym{CFG}{CFG}{Context-Free Grammar}
\newacronym{CNN}{CNN}{Convolutional Neural Network}
\newacronym{DSGE}{DSGE}{Dynamic Structured Grammatical Evolution}
\newacronym{GPU}{GPU}{Graphics Processing Unit}
\newacronym{ES}{ES}{Evolutionary Strategy}

\AtBeginEnvironment{align}{\setcounter{equation}{0}}

\newcommand{\veryshortarrow}[1][5pt]{\mathrel{%
   \vcenter{\hbox{\rule[-.5\fontdimen8\textfont3]{#1}{\fontdimen8\textfont3}}}%
   \mkern-4mu\hbox{\usefont{U}{lasy}{m}{n}\symbol{41}}}}

\begin{document}

\mainmatter  

\title{Incremental Evolution and Development of \\ Deep Artificial Neural Networks}

\author{Filipe Assun\c{c}\~ao\inst{1,2} \and Nuno Louren\c{c}o\inst{1} \and \\ Bernardete Ribeiro\inst{1} \and Penousal Machado\inst{1}}

\authorrunning{Assun\c{c}\~ao et al.}
\titlerunning{Incremental Evolution and Development of Deep Artificial Neural Networks}

\institute{CISUC, Department of Informatics Engineering,\\ University of Coimbra, Coimbra, Portugal\\
\email{\{fga, naml, bribeiro, machado\}@dei.uc.pt} \and LASIGE, Department of Informatics, Faculdade de Ciencias,\\ Universidade de Lisboa, Lisboa, Portugal}

\maketitle

\begin{abstract}
NeuroEvolution (NE) methods are known for applying Evolutionary Computation to the optimisation of Artificial Neural Networks (ANNs). Despite aiding non-expert users to design and train ANNs, the vast majority of NE approaches disregard the knowledge that is gathered when solving other tasks, i.e., evolution starts from scratch for each problem, ultimately delaying the evolutionary process. To overcome this drawback, we extend Fast Deep Evolutionary Network Structured Representation (Fast-DENSER) to incremental development. We hypothesise that by transferring the knowledge gained from previous tasks we can attain superior results and speedup evolution. The results show that the average performance of the models generated by incremental development is statistically superior to the non-incremental average performance. In case the number of evaluations performed by incremental development is smaller than the performed by non-incremental development the attained results are similar in performance, which indicates that incremental development speeds up evolution. Lastly, the models generated using incremental development generalise better, and thus, without further evolution, report a superior performance on unseen problems.
\keywords{Incremental Development, NeuroEvolution, Convolutional Neural Networks}
\end{abstract}

\glsresetall

\section{Introduction}
\label{sec:intro}

\gls{AutoML} is a sub-field of \gls{AI} that automates with little or no human-intervention the application of \gls{ML} approaches to the user's problem, avoiding the need for the manual tuning of the data pre-processing, the design and extraction of features, and/or the selection and parameterisation of the most suitable \gls{ML} model. The current work focuses on a branch of AutoML: \gls{NE}~\cite{Floreano2008}. NE applies \gls{EC} to search for \glspl{ANN}, enabling the optimisation of their structure (e.g., number of neurons, layers, connectivity), and/or learning (i.e., weights, or learning algorithm and its parameters). In other words, the ultimate goal of \gls{NE} is to empower non-expert \gls{ML} users with the ability to design effective \glspl{ANN}. 

One of the main limitations of \gls{NE} lies in the fact that the majority of the methods only address a specific problem, i.e., the \glspl{ANN} are evolved for one task, and when there is the need to solve a new problem the entire search procedure is re-started from scratch. Therefore, the methods do not take advantage of any of the information available from addressing previous similar tasks. In addition, \gls{NE} approaches tend to evolve large populations of individuals that are continuously optimised throughout a usually large number of generations. The evaluation of a single \gls{ANN} is time-consuming, because it often requires the training of the networks with a defined (or evolved) learning strategy. Consequently, the search for effective \glspl{ANN} resorting to \gls{NE} tends to be slow. This problem is even more striking when optimising \glspl{DANN}.

In this work we extend \gls{Fast-DENSER}~\cite{assunccao2019fdenser} to incremental development, i.e., we transfer and re-use the knowledge acquired when optimising \glspl{DANN} (architectures and learning strategies) to previous problems, and cumulatively apply it to learn new classification tasks. The main contributions of this work are the following:
\begin{itemize}
    \item The extension of the \gls{Fast-DENSER} framework to incremental development;
    \item The demonstration that \glspl{DANN} evolved by incremental development statistically outperform the canonical approach;
    \item The indication that incremental development speeds up evolution. When given the same number of generations, incremental development surpasses the performance of the evolution from scratch. For the same level of performance fewer generations are necessary;
    \item The evidence that the method works as expected in terms of evolution, i.e., knowledge from previously solved problems is introduced in any stage;
    \item The conclusion that the \glspl{DANN} that are evolved by incremental development generalise better than those obtained by the non-incremental version. The performance of the incrementally generated \glspl{DANN} is superior to their independent evolution counterparts in previously addressed, and in yet unaddressed problems.
\end{itemize}

The remainder of the document is organised as follows. Section~\ref{sec:background_sota} surveys related works in the field of \gls{NE} applied to \glspl{DANN}, and incremental development; Section~\ref{sec:fast_denser} details \gls{Fast-DENSER}; Section~\ref{sec:incremental_development} introduces the extension of \gls{Fast-DENSER} to incremental development; Section~\ref{sec:experimentation} presents the experimental setup and results; and Section~\ref{sec:conclusions} draws conclusions and addresses future work.

\section{Related Work}
\label{sec:background_sota}

NeuroEvolution (NE) approaches are usually grouped according to the target of evolution, i.e., topology~\cite{Gruau:1996:CCE:1595536.1595547,DBLP:conf/icga/MillerTH89}, learning (i.e., weights, parameters, or learning policies)~\cite{whitley1989applying,DBLP:journals/jmlr/GomezSM08,stanley2009hypercube}, or the simultaneous evolution of the topology and learning~\cite{stanley2002evolving,DBLP:conf/gecco/TurnerM13}. Nonetheless, more recent efforts have been put towards the proposal of methods that deal with the optimisation \glspl{DANN}, and thus we feel that it is more intuitive to divide them into small-scale~\cite{whitley1989applying,stanley2002evolving} and large-scale~\cite{stanley2009hypercube,DBLP:journals/corr/MiikkulainenLMR17,suganuma2017genetic,DBLP:conf/icml/RealMSSSTLK17,assuncao2018gpem} NE. The current paper focuses on the latter; a complete survey can be found in~\cite{baldominos2019automated}.

The problem of most of the methods that target the evolution of DANNs is that, even aided by \glspl{GPU} they tend to take a lot of time to find effective models. For example, CoDeepNEAT~\cite{DBLP:journals/corr/MiikkulainenLMR17} trains on 100 \glspl{GPU}, and Real et al. use 450 \glspl{GPU} for 7 days to perform each run~\cite{real2018regularized}. \gls{Fast-DENSER} takes approximately 4.7 days with a single \gls{GPU} to perform each run, and that is the reason why we have selected \gls{Fast-DENSER} for the current paper. There are methods that are computationally cheaper, e.g., Lorenzo and Nalepa~\cite{DBLP:conf/gecco/LorenzoN18} take about 120 minutes to obtain results; however, the speedup is obtained at the cost of the performance of the model.

To speedup evolution some authors have investigated the use of transfer learning in \gls{NE}. The main goal of transfer learning is to make use of the knowledge acquired when solving previous tasks to facilitate the resolution of others, enhancing lifelong learning~\cite{DBLP:conf/nips/Thrun95}. One of the most recurrent ideas is that of using past knowledge to provide a better start than random seeding (e.g., \cite{DBLP:conf/icnc/TirumalaAR16,NIPS2018_8056}).

A key problem on transfer (and even multi-task) learning is the representation. Verbancsics and Stanley~\cite{DBLP:journals/jmlr/VerbancsicsS10} demonstrate that transfer learning is most effective when the representation is the same for the multiple problems that are to be addressed. That is one of the advantages of using a grammar-based \gls{NE} approach such as \gls{Fast-DENSER}: the grammar nature of the method makes passing from one task to the next one transparent, and requires no changes to the individuals' representation.

Whilst some transfer learning works seek to learn high-level features that are generalisable across multiple domains (e.g., \cite{long2015learning}), our objective is to port individuals to warm start evolution to another problem, and in theory help to reach high performing solutions in less time. An example of a similar work, but where a hand designed network is used is introduced by Ciresan et al.~\cite{DBLP:conf/ijcnn/CiresanMS12}, where there is the transfer of knowledge from Latin digits recognition to uppercase letters, and from Chinese characters to uppercase Latin letters.

\section{Fast-DENSER}
\label{sec:fast_denser}

\begin{figure}[t!]
    \centering
    \scriptsize
    \begin{align}
        {<}\text{features}{>} ::= & \, {<}\text{convolution}{>}  \, | \,  {<}\text{convolution}{>}\\
                   & \, | \, {<}\text{pooling}{>} \, | \,  {<}\text{pooling}{>} \\
                   & \, | \, {<}\text{dropout}{>} \, | \, {<}\text{batch-norm}{>} \\
        {<}\text{convolution}{>} ::= & \, \text{layer:conv} \, \text{[num-filters,int,1,32,256]} \, \text{[filter-shape,int,1,2,5]} \\
                    & \, \text{[stride,int,1,1,3]} \, {<}\text{padding}{>} \, {<}\text{activation}{>} \, {<}\text{bias}{>}\\
        {<}\text{batch-norm}{>} ::= & \text{layer:batch-norm}\\
       {<}\text{pooling}{>} ::= & \, {<}\text{pool-type}{>} \, \text{[kernel-size,int,1,2,5]} \\
                    & \, \text{[stride,int,1,1,3]} \, {<}\text{padding}{>} \\
        {<}\text{pool-type}{>} ::= & \, \text{layer:pool-avg} \, | \, \text{layer:pool-max}\\
        {<}\text{padding}{>} ::= & \, \text{padding:same} \, | \, \text{padding:valid}\\
        {<}\text{classification}{>} ::= & \, {<}\text{fully-connected}{>} \, | \,  {<}\text{dropout}{>} \\
        {<}\text{fully-connected}{>} ::= & \, \text{layer:fc} \, {<}\text{activation}{>} \\
                    & \, \text{[num-units,int,1,128,2048} \, {<}\text{bias}{>} \\
        {<}\text{dropout}{>} ::= & \text{layer:dropput} \, \text{[rate,float,1,0,0.7]} \\
        {<}\text{activation}{>} ::= & \, \text{act:linear} \, | \, \text{act:relu} \, | \, \text{act:sigmoid}\\
        {<}\text{bias}{>} ::= & \, \text{bias:True} \, | \, \text{bias:False}\\
        {<}\text{softmax}{>} ::= & \, \text{layer:fc} \, \text{act:softmax} \, \text{num-units:10} \, \text{bias:True}\\
        {<}\text{learning}{>} ::= & \, {<}\text{bp}{>} \, {<}\text{early-stop}{>} \, \text{[batch\_size,int,1,50,500]} \\
                   & \, | \, {<}\text{rmsprop}{>} \, {<}\text{early-stop}{>} \, \text{[batch\_size,int,1,50,500]}  \\
                   & \, | \, {<}\text{adam}{>} \, {<}\text{early-stop}{>} \, \text{[batch\_size,int,1,50,500]} \\
         {<}\text{bp}{>} ::= & \, \text{learning:gradient-descent} \, \text{[lr,float,1,0.0001,0.1]}  \\
                   & \, \text{[momentum,float,1,0.68,0.99]} \\
                   & \, \text{[decay,float,1,0.000001,0.001]} \, {<}\text{nesterov}{>} \\
         {<}\text{nesterov}{>} ::= & \, \text{nesterov:True} \, | \, \text{nesterov:False} \\
         {<}\text{adam}{>} ::= & \, \text{learning:adam} \, \text{[lr,float,1,0.0001,0.1]} \, \text{[beta1,float,1,0.5,1]} \\
                   & \, \text{[beta2,float,1,0.5,1]} \, \text{[decay,float,1,0.000001,0.001]} \\
         {<}\text{rmsprop}{>} ::= & \, \text{learning:rmsprop} \, \text{[lr,float,1,0.0001,0.1]} \\
                   & \, \text{[rho,float,1,0.5,1]} \, \text{[decay,float,1,0.000001,0.001]} \\
         {<}\text{early-stop}{>} ::= & \, \text{[early\_stop,int,1,5,20]}
    \end{align}
    \caption{CFG for the optimisation of the topology and learning strategy of CNNs.}
    \label{fig:cnn_grammar}
\end{figure}

\gls{Fast-DENSER}~\cite{assunccao2019fdenser} is an extension of \gls{DENSER}~\cite{assuncao2018gpem}: a general-purpose grammar-based \gls{NE} approach for optimising \glspl{DANN}. \gls{DENSER} can search for any type of \gls{DANN}, and the target of evolution is specified in a \gls{CFG}. An example of a \gls{CFG} for encoding \glspl{CNN} is provided in Figure~\ref{fig:cnn_grammar}. The typical structure of \glspl{CNN} divides the topology into two parts: (i) layers for feature extraction (convolutional and pooling, lines 1-3), and layers for classification (fully-connected, line 11). The grammar of Figure~\ref{fig:cnn_grammar} explores these layer types, and also regularisation layers (dropout and batch normalisation, lines 1-3, and 11). Furthermore, the grammar enables the optimisation of the learning strategy (learning, lines 18-20). The parameters of each evolutionary unit (in the current work layers or learning algorithms) are kept in the grammar, and can be integer (e.g., the filter shape in line 4), float (e.g., the momentum in line 22) or closed choice (e.g., the bias in line 16). The integer and float parameters are represented by a block with the format: [variable-name, variable-type, num-values, min-value, max-value].

In addition to the \gls{CFG} we need to define the macro-structure, that establishes the search space, and points directly to the grammar production rules. The macro-structure sets the sequence of evolutionary units that the individuals are allowed to use, and is encoded as a list of tuples, where each position indicates the non-terminal symbol (that establishes a one-to-one mapping to the grammar, and is used as starting symbol), and the minimum and maximum number of expansions for that non-terminal symbol. For example, for \glspl{CNN}, an example of a macro-structure is $[$(features, 1, 10), (classification, 1, 2), (softmax, 1, 1), (learning, 1, 1)$]$. This macro-structure allows for \glspl{CNN} with between 3 and 13 layers, and where the learning strategy is optimised.

\begin{figure}[t!]
    \centering
    \includegraphics[width=.74\linewidth]{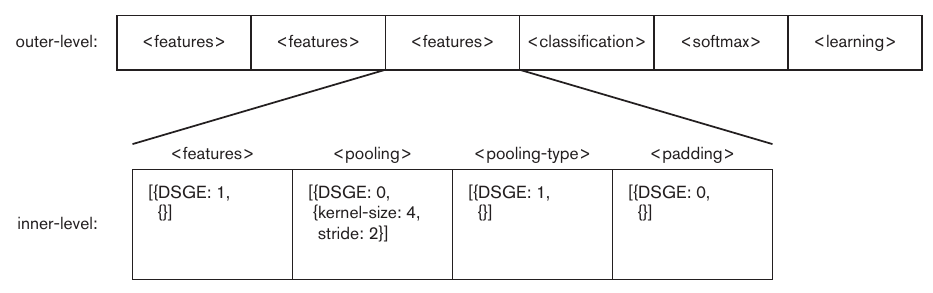}
    \caption{Example of the genotype of a candidate solution that encodes a CNN.}
    \label{fig:denser_genotype}
    \includegraphics[width=0.38\linewidth]{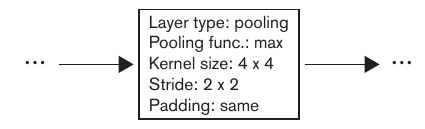}
    \caption{Phenotype of the layer specified by the inner-level of Figure~\ref{fig:denser_genotype}}
    \label{fig:denser_phenotype}
\end{figure}

The genotype of the candidate solutions is organised into two levels: (i) the outer-level encodes the sequence of evolutionary units (with respect to the macro-structure), and sets the non-terminal symbol that is used as initial symbol for the grammatical derivation; and (ii) the inner-level corresponds to each outer-level position and encodes the parameters of a specific evolutionary unit. The inner-level genotype is similar to the genotype of \gls{DSGE}; for more details on \gls{DSGE} refer to \cite{Lourenco2018}. An example of the genotype and corresponding phenotype of a candidate solution are represented in Figures~\ref{fig:denser_genotype} and~\ref{fig:denser_phenotype}, respectively.

The representation of the candidate solutions in \gls{DENSER} and \gls{Fast-DENSER} is the same. The differences between the two approaches lie in the evolution of the population and in the evaluation of the candidate solutions. In \gls{DENSER} evolution is conducted as in a standard Genetic Algorithm, where in each generation a large population of individuals is evaluated and offspring is generated. Contrary, \gls{Fast-DENSER} follows a (1+$\lambda$) \gls{ES}, and therefore in each generation fewer individuals are evaluated. The results have demonstrated that \gls{Fast-DENSER}, with the same individual evaluation scheme, can generate individuals that have the same quality as those generated by \gls{DENSER}, in a fraction of the time. More precisely, there is a speedup of 20x from \gls{DENSER} to \gls{Fast-DENSER}. In addition, \gls{Fast-DENSER} is extended to enable the generation of fully-trained \glspl{DANN}, i.e., networks that need no further training by the end of the evolutionary process. To this end, \gls{Fast-DENSER} evaluates the individuals for a maximum \gls{GPU} training time. However, the maximum training time granted to each individual can grow continuously as required. The networks that are likely to benefit from longer training cycles are given access to a greater evaluation time as evolution proceeds.

\section{Incremental Development of Deep Neural Networks}
\label{sec:incremental_development}

Experiments on previous work have shown that \gls{Fast-DENSER}, given the same computational time budget, can obtain results that are superior to those reported by \gls{DENSER}. The results are achieved without taking advantage of any of the knowledge acquired when solving other problems. In this paper we investigate the impact of building the networks incrementally, i.e., we take into account the \glspl{DANN} that are generated for solving previous related problems, speeding up evolution, and possibly finding more effective solutions.

\begin{figure}[t!]
    \centering
    \includegraphics[width=0.68\linewidth]{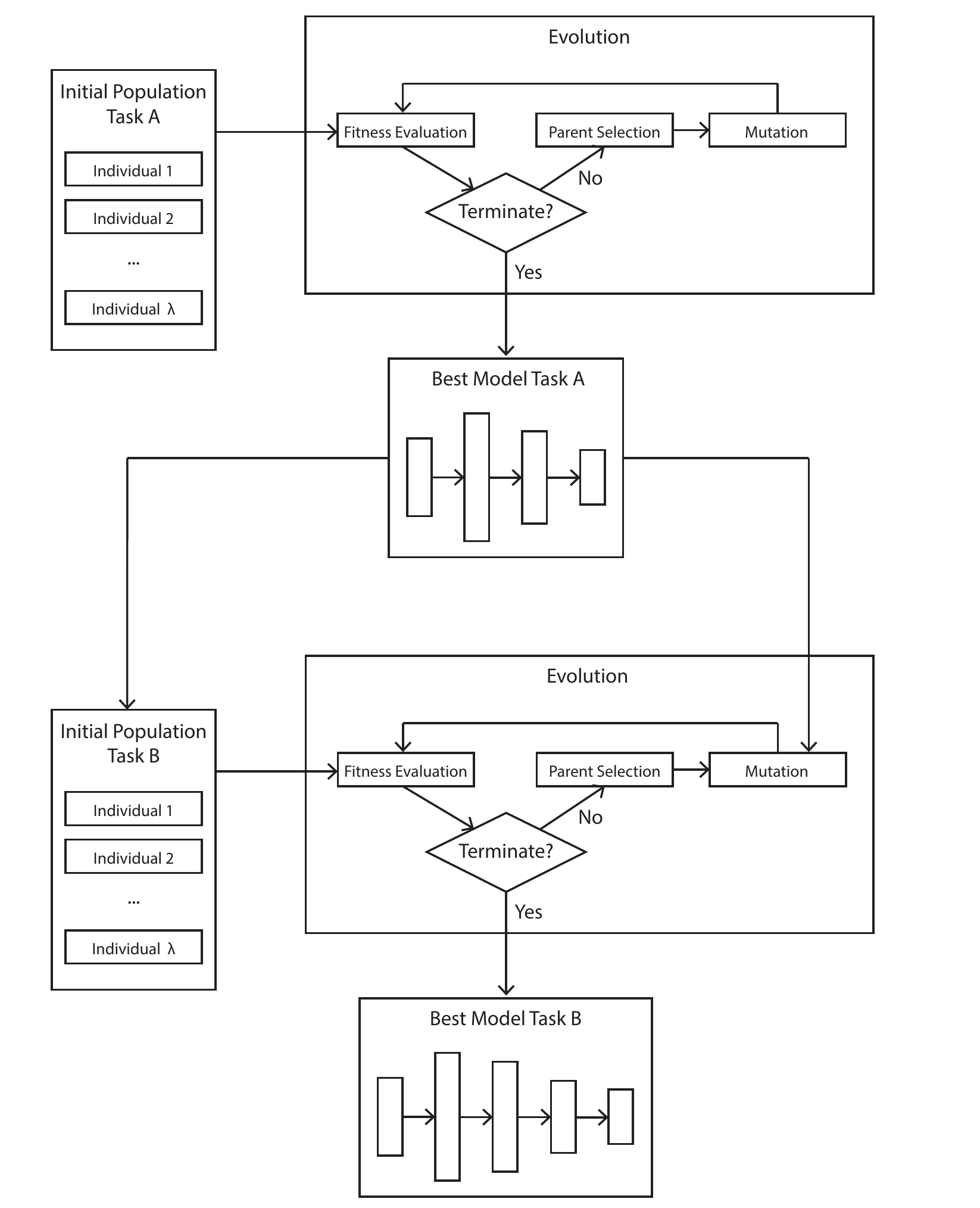}
    \caption{Incremental development Fast-DENSER flow-chart.}
    \label{fig:fast_denser_incremental}
\end{figure}

The flow-chart that illustrates the extension of \gls{Fast-DENSER} to incremental development is depicted in Figure~\ref{fig:fast_denser_incremental}. It shows how the method proceeds to address two different tasks A and B. For the first problem, task A, the method works similarly to \gls{Fast-DENSER}: an initial population is randomly created and evolves until the stop criterion is met. The difference occurs when we solve a new problem, given that we have information on a prior one. For task B, the creation of the initial population takes into account the best model found for a previous problem (in this case task A). During evolution the past knowledge can also be incorporated. This rationale is generalised for more than two problems, i.e., in case we later address a task C, we use the knowledge obtained when addressing tasks A and B. Next, we will discuss how the prior knowledge is introduced in the initial population, and during evolution. 

The initial population is formed by individuals that can be either entirely generated at random or that can use sets of evolutionary units from past models. The evolutionary units are transferred taking into account the macro-structure. For example, considering the macro-structure introduced above for \glspl{CNN}, $[$(features, 1, 10), (classification, 1, 2), (softmax, 1, 1), (learning, 1, 1)$]$, the initial population can contain individuals that (i) have all the layers comprising the feature extraction, and generate the classification layers at random; (ii) generate at random the feature extraction layers, and copy the layers that perform classification from previous models; (iii) copy only the learning evolutionary unit, and generate all the remaining ones at random; (iv) generate all evolutionary units at random, not using any previous knowledge; or (v) any other possible combination. It is important to mention that this incremental development approach only focuses on the evolutionary units, and consequently the weights are not transferred from previous models. At most we allow the learning strategy (which is an evolutionary unit) to be ported.

The models generated for solving each of the previously addressed problems are also important during evolution. The mutations in \gls{Fast-DENSER} are tailored for manipulating \glspl{DANN}: they enable the addition, removal, and/or duplication of any evolutionary unit, and the perturbation of the integer and/or float values. The duplication mutation, as the name suggests, replicates a given evolutionary unit by reference, and thus, any mutation that later affects this evolutionary unit changes all of its copies. In the incremental development version of \gls{Fast-DENSER} the duplication can copy evolutionary units either from the individual or from any of the best models that were generated for solving previous tasks.

The individuals are evaluated only on the new problem. Therefore, up to the moment, this method is incremental in the sense that the \glspl{DANN} for solving new and unseen problems do not kick off evolution from scratch. The incremental development does not mean that by the end of evolution the generated models can solve multiple tasks. However, it is expected that the models that are built considering previous knowledge generalise better than those that are always evolved from a random population. That is, we expect the models generated by incremental development to perform well in other tasks when re-trained.

\section{Experimentation}
\label{sec:experimentation}

To compare the incremental and non-incremental versions of \gls{Fast-DENSER} we consider four computer vision datasets: MNIST, SVHN, Fashion-MNIST, and CIFAR-10 (summarised in Section~\ref{sec:datasets}). In particular, we conduct experiments for the following setups: (i) MNIST; (ii) SVHN; (iii) Fashion-MNIST; (iv) CIFAR-10; (v) MNIST $\rightarrow$ SVHN; (vi) MNIST $\rightarrow$ SVHN $\rightarrow$ Fashion-MNIST; (vii) MNIST $\rightarrow$ SVHN $\rightarrow$ CIFAR-10. The symbol $\rightarrow$ denotes the incremental build of the model from one task to the next. The setups are chosen according to the relatedness and expected difficulty of the tasks: the MNIST and SVHN datasets are composed by digits, and then transferred to two different domains, Fashion-MNIST, and CIFAR-10. The parameters required for the conducted experiments are detailed in Section~\ref{sec:exp_setup}. The experimental results are divided into three sections. First, in Section~\ref{sec:results_incremental_development} we analyse the evolutionary performance when evolving \glspl{DANN} for MNIST, SVHN, Fashion-MNIST, and CIFAR-10 with and without incremental development. Second, in Section~\ref{sec:results_incremental_development_top} we investigate the incremental development of the topologies. Third, in Section~\ref{sec:results_robustness}, we analyse the generalisation ability of the different models. The experimental results are discussed in Section~\ref{sec:exp_discussion}.

\subsection{Datasets}
\label{sec:datasets}

\begin{table}[t!]
    \centering
    \begin{tabular}{c|c|c|c|c}
        \textbf{Dataset} & \textbf{Train Set Size} & \textbf{Test Set Size} & \textbf{Number of Classes} & \textbf{Shape} \\ \hline
        MNIST & 60000 & 10000 & 10 & 28$\times$28$\times$1 \\
        SVHN & 73257 & 26032 & 10 & 32$\times$32$\times$3 \\
        Fashion-MNIST & 60000 & 10000 & 10 & 28$\times$28$\times$1\\
        CIFAR-10 & 50000 & 10000 & 10 &  32$\times$32$\times$3\\
    \end{tabular}
    \vspace{2.5pt}
    \caption{Description of the datasets.}
    \label{tab:datasets}
\end{table}

The experiments are conducted in 4 datasets: MNIST, SVHN, Fashion-MNIST, and CIFAR-10. The characteristics of the datasets are summarised in Table~\ref{tab:datasets}. The shape of the instances is formatted as width $\times$ height $\times$ number of channels; grayscale images have one channel, and RGB images have three channels.  A brief overview of the dataset instances is provided next.
\begin{description}
    \item[MNIST]~\cite{mnist} -- handwritten digits from 0 to 9. The instances are pre-processed: size-normalized, and centered;
    \item[SVHN]~\cite{svhn} -- digits gathered from real-world images from house numbers in Google Street View images;
    \item[Fashion-MNIST]~\cite{fashion-mnist} -- similar to MNIST, where the images of handwritten digits are replaced by fashion clothing items: top, trouser, pullover, dress, coat, sandal, shirt, sneaker, bag, and ankle boot;
    \item[CIFAR-10]~\cite{cifar10} -- real-world pictures of objects that are of one of the following classes: airplane, automobile, bird, cat, deer, dog, frog, horse, ship, and truck. 
\end{description}

\subsection{Experimental Setup}
\label{sec:exp_setup}

\begin{table}[t!]
    \centering
    \begin{tabular}{c | c}
        \textbf{Evolutionary Engine Parameter} & \textbf{Value} \\ \hline
        Number of runs & 10 \\ 
        Number of generations & 20 / 30 / 50 / 100 \\  
        $\lambda$ & 5 \\
        Add layer rate & 25\% \\
        Duplicate layer rate & 15\%\\
        Remove layer rate & 25\% \\
        Grammatical mutation rate & 15\% \\ \hline
        \textbf{Dataset Parameter} & \textbf{Value} \\ \hline
        Evolutionary Validation set & 3500 instances \\
        Evolutionary Test set & 3500 instances \\ 
        Evolutionary Train set &  Remaining instances \\ 
        Test set & Check Table~\ref{tab:datasets} \\ \hline
        \textbf{Training Parameter} & \textbf{Value} \\ \hline
        Training Time & 10 min. \\
        Loss & Categorical Cross-entropy \\ \hline
        \textbf{Data Augmentation Parameter} & \textbf{Value} \\ \hline
        Padding & 4 \\
        Random crop & 4 \\
        Horizontal flipping & 50\% \\
    \end{tabular}
    \vspace{2.5pt}
    \caption{Experimental parameters.}
    \label{tab:exp_parameters}
\end{table}

The experimental parameters are detailed in Table~\ref{tab:exp_parameters}. The table is organised into 4 sections: (i) evolutionary engine -- parameters related to \gls{Fast-DENSER} (1+$\lambda$)-\gls{ES}; (ii) dataset -- parameters concerned with the dataset partitioning; (iii) training -- parameters associated with the training of the \glspl{DANN}; and (iv) data augmentation -- parameters required for the dataset augmentation strategy.

The number of generations is different for each of the datasets. In particular, we perform 20, 30, 50, and 100 generations for the MNIST, Fashion-MNIST, SVHN, and CIFAR-10 datasets, respectively. The number of generations for each dataset was set empirically based on previous experiments, and according to how challenging each problem is expected to be. The grammatical mutation rate is a \gls{DSGE} parameter that stands for the probability of changing any of the grammar expansion possibilities or integer/float parameter values.

The dataset section of Table~\ref{tab:exp_parameters} defines how we partition the train set of each dataset, i.e., for each run the train set is divided (in a stratified way) into three independent folds: (i) evolutionary train -- used for training the \glspl{DANN}; (ii) evolutionary validation -- used to perform early stop; and (iii) evolutionary test -- used to measure the fitness of the individual, which is measured using the accuracy. The test set (that is different from the evolutionary test set) is kept out of evolution and is used only after the end of the evolutionary search. It measures how well the models behave beyond the data used during evolution, and enables the unbiased evaluation of the performance.  

The datasets, as discussed in Section~\ref{sec:datasets}, have different shapes: MNIST and Fashion-MNIST are 28$\times$28$\times$1, and SVHN and CIFAR-10 are 32$\times$32$\times$3. To facilitate the application of the optimised \glspl{DANN} to all datasets we reshape the MNIST and Fashion-MNIST to 32$\times$32$\times$3. The image width and height are resized using the nearest neighbour method, and to pass from one to three channels we replicate the single channel three times. All the datasets are applied the same data augmentation strategy: padding, random cropping, horizontal flipping, and re-scaling to $[0,1]$. We do not subtract the mean image nor normalize. 

The networks are trained for an initial maximum \gls{GPU} time of 10 minutes, and thus it is important to mention that we are performing each evolutionary run in a GeForce GTX 1080 Ti \gls{GPU}. For the experiments conducted on this paper we use the grammar of Figure~\ref{fig:cnn_grammar}, and the macro-structure: $[$(features, 1, 30), (classification, 1, 10), (softmax, 1, 1), (learning, 1, 1)$]$. The code for \gls{Fast-DENSER} can be found in the GitHub repository \url{https://github.com/fillassuncao/fast-denser3}.

\subsection{Experimental Results: Incremental Development}
\label{sec:results_incremental_development}

\begin{table}[t!]
    \centering
    \begin{tabular}{c|c|c}
        \textbf{Dataset} & \textbf{Evolutionary Accuracy}& \textbf{Test Accuracy} \\ \hline
        MNIST            & 98.86 $\pm$ 0.465             & 98.80 $\pm$ 0.298      \\ \hdashline
        SVHN             &  93.28 $\pm$ 0.863             &  93.31 $\pm$ 0.955      \\
        MNIST $\veryshortarrow$ SVHN & \bf 94.01 $\pm$ 0.891 & \bf 94.04 $\pm$ 0.887      \\ \hdashline
        Fashion          &  92.42 $\pm$ 1.224             &  91.41 $\pm$ 1.049      \\
        MNIST $\veryshortarrow$ SVHN $\veryshortarrow$ Fashion & \bf 93.92 $\pm$ 0.930 & \bf 92.96 $\pm$ 0.742 \\ \hdashline
        CIFAR-10         & 87.18 $\pm$ 1.242             & 86.19 $\pm$ 1.672      \\
        MNIST $\veryshortarrow$ SVHN $\veryshortarrow$ CIFAR-10 & \bf 89.06 $\pm$ 1.488& \bf 88.19 $\pm$ 1.669 \\
    \end{tabular}
    \vspace{2.5pt}
    \caption{Average performance of the optimised DANNs. The results are averages of 10 independent runs. Bold marks the highest average performance values.}
    \label{tab:evolutionary_results_average}
\end{table}

We start by comparing the \glspl{DANN} generated by \gls{Fast-DENSER} with and without incremental development in terms of performance. The results are summarised in Table~\ref{tab:evolutionary_results_average}. We report the evolutionary accuracy (i.e., fitness), and the test accuracy (i.e., the accuracy of the models on an unseen partition of the datasets). The results are averages of 10 independent runs. The first conclusion is that given the same computational time (number of generations), the results reported by the incremental development are always superior to those of when evolution starts from scratch. In particular, the performance of MNIST $\veryshortarrow$ SVHN is superior to the performance of SVHN, the performance of MNIST $\veryshortarrow$ SVHN $\veryshortarrow$ Fashion is superior to the performance of Fashion, and the performance of MNIST $\veryshortarrow$ SVHN $\veryshortarrow$ CIFAR-10 is superior to the performance of CIFAR-10. 



To better acknowledge the differences between the fittest \glspl{DANN} generated with and without incremental development we use statistical tests. To check if the samples follow a Normal Distribution we use the Kolmogorov-Smirnov and Shapiro-Wilk tests, with $\alpha = 0.05$. The tests reveal that the data does not follow any distribution and thus the non-parametric Mann-Whitney U test ($\alpha$ = 0.05) is used to perform the comparisons between the setups. The statistical tests show that the results of MNIST $\veryshortarrow$ SVHN $\veryshortarrow$ Fashion, and MNIST $\veryshortarrow$ SVHN $\veryshortarrow$ CIFAR-10 are statistically superior (in evolution and test) to Fashion (evolutionary p-value=0.00736, test p-value=0.00278), and CIFAR-10 (evolutionary p-value=0.00804, test p-value=0.01552), respectively. The effect size is large for all the statistically significant comparisons (r $> $0.5). The difference between MNIST $\veryshortarrow$ SVHN and SVHN is not statistically significant (evolutionary p-value=0.05876, test p-value=0.0536). With only 20 generations the MNIST setup is the one that attains the highest average accuracy results. This indicates that it is an easy to solve problem and consequently no much knowledge is acquired from addressing it. This is a well-known fact: a simple fully-connected network is able to attain good performances in the MNIST dataset.

The above results prove that incremental development, given the same number of generations, designs \glspl{DANN} that outperform those generated without incremental development. On the other hand, what happens when, for each setup, we only let evolution to be conducted for a smaller amount of generations, so that the cumulative number of generations is not superior to that of when evolution is conducted from scratch? The cumulative number of generations is the sum of the number of generations of each incremental step. For example, for the MNIST $\veryshortarrow$ SVHN, the cumulative number of generations is 70 ($20 + 50$). In this scenario we consider 30, 0, and 30 generations for the MNIST $\veryshortarrow$ SVHN, MNIST $\veryshortarrow$ SVHN $\veryshortarrow$ Fashion, and MNIST $\veryshortarrow$ SVHN $\veryshortarrow$ CIFAR-10 setups, respectively. The average evolutionary performance of the 10 fittest networks slightly decreases to 93.69 $\pm$ 0.912, 92.91 $\pm$ 1.15, and 87.13 $\pm$ 2.225, respectively for the MNIST $\veryshortarrow$ SVHN, MNIST $\veryshortarrow$ SVHN $\veryshortarrow$ Fashion, and MNIST $\veryshortarrow$ SVHN $\veryshortarrow$ CIFAR-10 setups. With these results there is no statistical difference for any of the setups, i.e., with incremental development, given a cumulative search time that equals the search time from scratch, we are able to generate \glspl{DANN} that report the same performance as those optimised without incremental development for more generations. In other words, the use of previous knowledge speeds up evolution.


\begin{table}[t!]
    \centering
    \begin{tabular}{c|c|c}
        \textbf{Dataset} & \textbf{Evolutionary Accuracy} & \textbf{Test Accuracy}\\ \hline
        MNIST & 99.46 & 99.12 \\ \hdashline
        SVHN & 94.20 & 93.88\\
        MNIST $\veryshortarrow$ SVHN & \bf 94.80 & \bf 94.14 \\ \hdashline
        Fashion & 93.91 & 92.92 \\
        MNIST $\veryshortarrow$ SVHN $\veryshortarrow$ Fashion & \bf 94.80 & \bf 93.92 \\ \hdashline
        CIFAR-10 & 88.74 & 88.14 \\
        MNIST $\veryshortarrow$ SVHN $\veryshortarrow$ CIFAR-10 & \bf 91.06 & \bf 89.79 \\
    \end{tabular}
    \vspace{2.5pt}
    \caption{Accuracy of the best performing DANN for each of the setups. Bold marks the highest performance value.}
    \label{tab:evolutionary_results_best}
\end{table}

In addition to analysing the average performance over the 10 evolutionary runs we also focus on the overall best found \gls{DANN}, i.e., the fittest \gls{DANN} among the conducted runs. This analysis is important considering that in a real-world scenario by the end of evolution what really interests the user is the best found model, which is the one to potentially be deployed live. To avoid an unbiased choice of the best model for each dataset, the decision is taken only with regard to the evolutionary performance. The results are reported in Table~\ref{tab:evolutionary_results_best}, and once again show that the best results are obtained by incremental development. The most striking result is the one of CIFAR-10, where the difference introduced by incremental development is the highest.

\subsection{Experimental Results: Topology Analysis}
\label{sec:results_incremental_development_top}

\begin{figure}[t!]
    \centering
    \includegraphics[width=.68\linewidth]{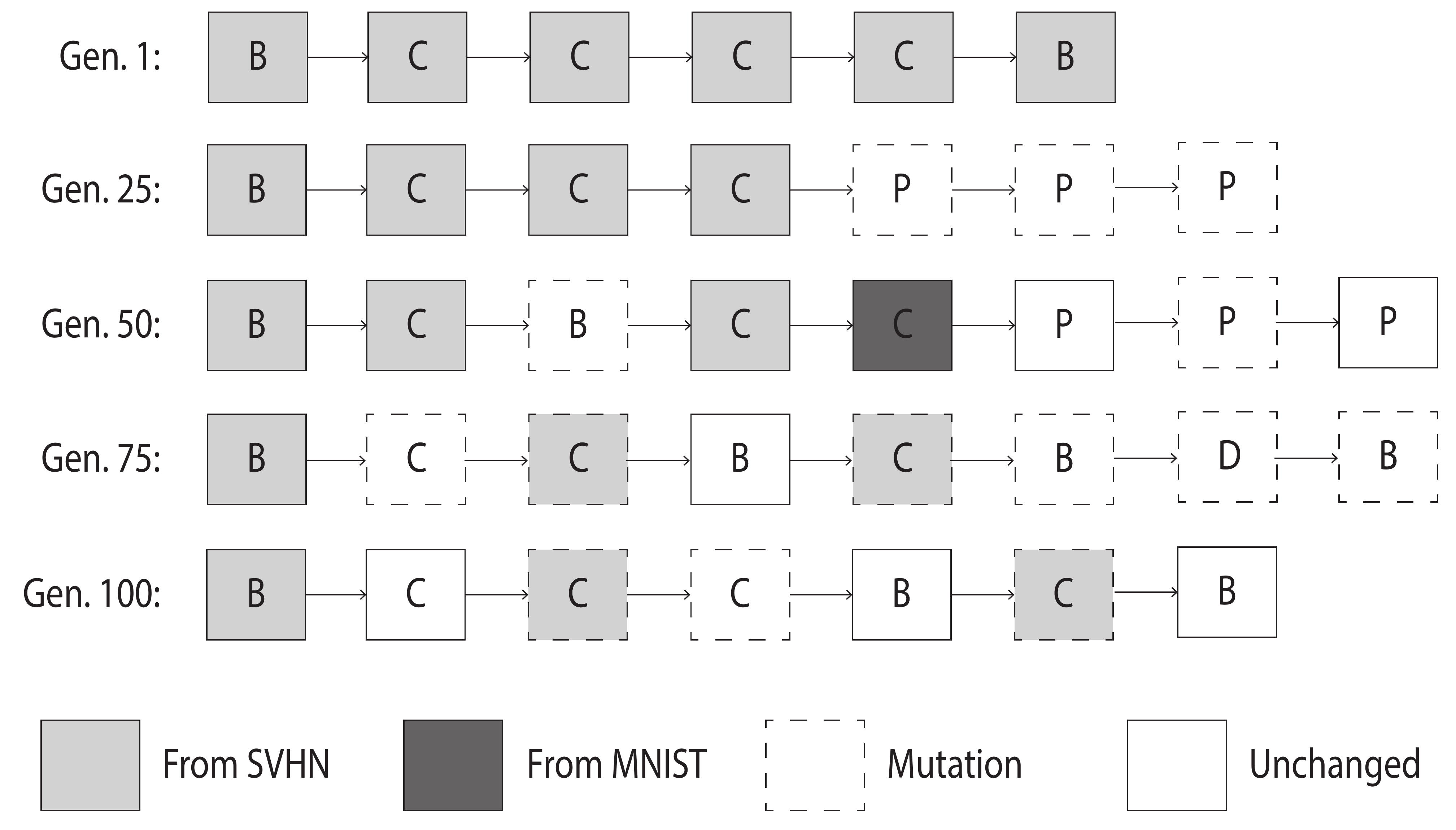}
    \caption{Overview of the evolution on the incremental development setup MNIST $\veryshortarrow$ SVHN $\veryshortarrow$ CIFAR-10. We provide a snapshot of the feature-layers of the best individual on the 1st, 25th, 50th, 75th, and 100th generations. For space constraints we focus on the feature extraction layers: Convolutional (C), Pooling (P), Batch-Normalization (B), and Dropout (D).}
    \label{fig:snapshot}
\end{figure}

To analyse the behaviour of incremental development from a structural point of view we inspect the topology of the best networks as evolution proceeds. Figure~\ref{fig:snapshot} shows the evolution of the structure of the networks on the setup MNIST $\veryshortarrow$ SVHN $\veryshortarrow$ CIFAR-10. Because of space constraints we select the setup where more generations were performed, and present the snapshots of the run that generates the \gls{DANN} with the median fitness value, i.e., we order the runs according to the fitness of the best generated \gls{DANN} and select the 6th run. We choose the median run to avoid a biased selection over the worst or best results. We focus only on the feature extraction layers. The figure's goal is to illustrate the exploration of knowledge incorporation, and thus the parameters of the layers are omitted.

The figure makes it evident that the amount of layers that come from previously addressed tasks without any change diminishes as evolution proceeds. That is the expected behaviour: in the initial generation the fittest \gls{DANN} re-uses all layers from the best network generated to address the SVHN, and across generations these layers are adapted to tackle the CIFAR-10 (e.g., convolutional in generation 75). During evolution new layers are also randomly created (e.g., batch-normalization in generation 50), and others removed (e.g., dropout in generation 100). Similarly to the the non-incremental approach, new random layers can be added, but in addition, in the incremental development strategy we can also add layers that come from the previously solved tasks (e.g., convolutional layer that is transferred from the MNIST in generation 50). 

The snapshots prove that incremental development is able to generate better results based on the re-use of evolutionary units that aid solving previous problems. The evolutionary units are not only incorporated in the generation of the initial population, but also during evolution. We also inspect the evolutionary results of other setups and the conclusions are inline with the reported.

\subsection{Experimental Results: Generalisation of the Models}
\label{sec:results_robustness}

With the objective of studying the generalisation ability of the generated models we measure their performance on all the considered datasets. For example, we take the best generated solutions for the MNIST dataset and apply them to the SVHN, Fashion and CIFAR-10 datasets without further evolutionary optimisation. The networks are re-trained on the target datasets with the same topology and learning strategy that is optimised for the source task. Table~\ref{tab:robustness} summarises the test results for all the setups. The values in bold mark the best generalisation performance, i.e., the best performance of the setup (row) that has not yet seen the dataset (column), e.g., for the CIFAR-10 dataset (last column), except for the setups that specifically target this dataset (CIFAR-10, and MNIST$\veryshortarrow$SVHN$\veryshortarrow$CIFAR-10), the setup that attains the highest performance is MNIST$\veryshortarrow$SVHN$\veryshortarrow$Fashion, and thus this is the setup that is marked in bold. 

The analysis of the results shows that incremental development always generates better results, even for tasks that have not been addressed previously. To better understand the differences we perform a statistical analysis, and compare the performances reported by the non-incremental and incremental approaches. Therefore we compare the SVHN and MNIST$\veryshortarrow$SVHN setups on the MNIST, Fashion, and CIFAR-10 datasets, and we do similarly with the remaining pairs: Fashion vs. MNIST$\veryshortarrow$SVHN$\veryshortarrow$Fashion, and CIFAR-10 vs. MNIST$\veryshortarrow$SVHN$\veryshortarrow$CIFAR-10. The same conditions of the above statistical comparison are applied. The statistical tests reveal that there are only significant differences between the Fashion, and MNIST$\veryshortarrow$SVHN$\veryshortarrow$Fashion setups, with p-values of 0.02574, and 0.01732, respectively for the SVHN and CIFAR-10 datasets (the effect size is large). The direct comparison for the dataset used for evolution (in this case Fashion) was performed above and revealed a statistical significance in favour of incremental development for the setups that include two incremental development steps. 

In case we order the datasets by difficulty, given by the non-incremental test performance on each dataset, we have MNIST, SVHN, Fashion, and CIFAR-10, where the leftmost is the easiest one, and the rightmost is the most challenging. From these results we hypothesise that superior generalisation performances are obtained by incremental development, when passing from more simple to more challenging datasets. That is the reason why there is no statistical difference in the CIFAR-10 vs. MNIST$\veryshortarrow$SVHN$\veryshortarrow$CIFAR-10 setups: the CIFAR-10 is per-se more challenging to solve than the remaining ones, and therefore, as already noticed in a previous article~\cite{assuncao2018gpem}, the \glspl{DANN} generated for addressing CIFAR-10 tend to be able to solve other easier problems. The remarkable aspect of incremental development is when a \gls{DANN} optimised for Fashion is able to get better results on the CIFAR-10, compared to when the \glspl{DANN} for Fashion are not evolved in an incremental fashion.

\begin{table}[t!]
    \centering
    \footnotesize 
    \begin{tabular}{c|c|c|c|c}
                                                         & MNIST              & SVHN               & Fashion            & CIFAR-10           \\ \hline
    MNIST                                                & 98.80$\pm$0.298  & 71.31$\pm$29.60  & 90.17$\pm$1.842  & 63.63$\pm$23.29  \\ \hdashline
    SVHN                                                 & 96.87$\pm$5.426  & 93.31$\pm$0.955  & 91.60$\pm$1.289  & 78.49$\pm$7.899  \\ 
    MNIST$\veryshortarrow$SVHN                           & 98.93$\pm$0.266  & 94.04$\pm$0.887  & 91.83$\pm$1.312  & 82.58$\pm$2.414   \\ \hdashline
    Fashion                                              & 92.73$\pm$16.75  & 89.16$\pm$3.551  & 91.41$\pm$1.049  & 77.32$\pm$4.893  \\ 
    MNIST$\veryshortarrow$SVHN$\veryshortarrow$Fashion   & 98.89$\pm$0.273  & \textbf{92.48$\pm$2.167}  & 92.96$\pm$0.742  & \textbf{83.47$\pm$2.294}  \\ \hdashline
    CIFAR-10                                             & 99.06$\pm$0.039  & 90.18$\pm$9.282  & 92.91$\pm$0.479  & 86.19$\pm$1.672  \\ 
    MNIST$\veryshortarrow$SVHN$\veryshortarrow$CIFAR-10  & \bf 99.11$\pm$0.071  & 90.08$\pm$5.924  & \bf 93.16$\pm$0.3328  & 88.19$\pm$1.669  \\
    \end{tabular}
    \vspace{2.5pt}
    \caption{Performance of the evolved DANNs when applied to other datasets. The results are averages of 10 independent runs, where each \gls{DANN} is trained 5 times. The setups are the table rows, and the datasets the columns.}
    \label{tab:robustness}
\end{table}





\subsection{Discussion}
\label{sec:exp_discussion}

The results presented in the previous sections compare in terms of performance, topology, and generalisation ability the search conducted by non-incremental and incremental development. The evolutionary results show that given the same search time the \glspl{DANN} obtained by incremental development statistically outperform the non-incremental counterparts. On the other hand, the incremental strategy speeds up evolution, and given the same cumulative search time reports results that match the non-incremental performances. 

The speedup in evolution is facilitated by the warm-start of incremental development, and possibility to still incorporate knowledge from previous tasks by mutation as generations proceed. We show an example of this by representing several snapshots of a network across generations. In particular for the selected run of the MNIST$\veryshortarrow$SVHN$\veryshortarrow$CIFAR-10 setup, on the first generation the best individual replicates all the layers from the MNIST$\veryshortarrow$SVHN setup, which are continuously modified and adapted to the CIFAR-10. During evolution the parameters of the layers that are copied from the previous setup are changed, new layers (random, and from previous setups) are added, and others removed. That is, the behaviour of the incremental development evolution is the expected. 

Finally, we analyse the generalisation ability of the generated \glspl{DANN}. Without further evolution, i.e., with the same topology and learning strategy obtained when optimising a \gls{DANN} for a specific task, we re-train the \glspl{DANN} on the remaining datasets. The results show that, on average, the incremental development results are superior to the non-incremental results. Moreover, the results are statistically significant when the generated \glspl{DANN} are applied to a more difficult task than that where they were generated. This indicates that incremental development helps in learning increasingly more challenging tasks, and that there are not major differences when performing the opposite.

\section{Conclusions and Future Work}
\label{sec:conclusions}

Motivated by the difficulty and burden in the design of \glspl{DANN} we investigate how to incorporate past knowledge to aid evolution. In particular, we extend \gls{Fast-DENSER} -- a general-purpose grammar-based \gls{NE} framework -- to take advantage of the evolutionary units acquired when optimising \glspl{DANN} for previous tasks. This novel incremental developmental approach enables the incorporation of knowledge from any of the previously addressed tasks in any stage of evolution: both during the generation of the initial population, and by the application of mutations, as the generations proceed.

The results prove that incremental development improves the search performed by \gls{Fast-DENSER} enabling it to obtain statistically superior results. Additionally, incremental development speeds up evolution, being able to obtain the same results as non-incremental evolution given the same cumulative search time, i.e., less generations are used for the target dataset. In addition, the \glspl{DANN} obtained by the end of evolution generalise better when we use incremental development in the search: the networks designed for easy problems perform better in more challenging and yet unseen tasks. 

The future work will target three different directions: (i) apply the incremental development methodology to a wider set of tasks and domains; (ii) extend the approach to modular evolution; and (iii) seek ways to transfer not only the evolutionary unit but also the weights (in case the evolutionary units are layers).

\section*{Acknowledgments}

\noindent This work is partially funded by: Funda\c{c}\~ao para a Ci\^encia e Tecnologia (FCT), Portugal, under the PhD grant SFRH/BD/114865/2016, and the project grant DSAIPA/DS/0022/2018 (GADgET). We also thank the NVIDIA Corporation for the hardware granted to this research.

\bibliography{bibliography}
\bibliographystyle{splncs03}

\end{document}